\documentclass{article}

\usepackage{microtype}
\usepackage{graphicx}
\usepackage{subfigure}
\usepackage{booktabs} 

\usepackage{amsmath}
\usepackage{amssymb}
\usepackage{mathtools}
\usepackage{amsthm}
\usepackage[section]{placeins}  

\usepackage{hyperref}

\usepackage[final]{sods_2023}

\usepackage{amsmath}
\usepackage{amssymb}
\usepackage{mathtools}
\usepackage{amsthm}

\usepackage[capitalize,noabbrev]{cleveref}

\theoremstyle{plain}

\theoremstyle{definition}

\theoremstyle{remark}

\usepackage[textsize=tiny]{todonotes}

\title{Complex Preferences for Different Convergent Priors in Discrete Graph Diffusion}

\author{
Alex M. Tseng\\
\texttt{tseng.alex@gene.com}\\
\And
Nathaniel Diamant\\
\texttt{diamant.nathaniel@gene.com}\\
\And Tommaso Biancalani\\
\texttt{biancalt@gene.com}\\
\And
\And Gabriele Scalia\\
\texttt{scalia.gabriele@gene.com}\\
\\
Department of Artificial Intelligence and Machine Learning\\
Research and Early Development\\
Genentech
}

\begin{document}

\maketitle

\begin{abstract}
Diffusion models have achieved state-of-the-art performance in generating many different kinds of data, including images, text, and videos. Despite their success, there has been limited research on how the underlying diffusion process and the final convergent prior can affect generative performance; this research has also been limited to continuous data types and a score-based diffusion framework. To fill this gap, we explore how different \textit{discrete} diffusion kernels (which converge to different prior distributions) affect the performance of diffusion models for graphs. To this end, we developed a novel formulation of a \textit{family} of discrete diffusion kernels which are easily adjustable to converge to different Bernoulli priors, and we study the effect of these different kernels on generative performance. We show that the quality of generated graphs is sensitive to the prior used, and that the optimal choice cannot be explained by obvious statistics or metrics, which challenges the intuitions which previous works have suggested.
\end{abstract}

\section{Introduction}
In recent years, diffusion models have been applied successfully to many different problems and data types, achieving state-of-the-art generation quality \cite{Sohl-Dickstein2015,Ho2020,Song2021,Dhariwal2021,Rombach2021}. Despite how central the underlying diffusion process is to a diffusion model, however, there has been very limited research that explores how different diffusion processes affect generative performance. A few works have found that performance \textit{can} be affected by the diffusion process, but these findings have largely been limited to diffusion on continuous objects, and where diffusion time is also continuous (i.e. using a stochastic-differential-equation framework) \cite{Song2021,Dockhorn2021,Karras2022}. In contrast, \emph{discrete diffusion models} \cite{Austin2021} have recently emerged as a more effective way to model intrinsically discrete objects, such as graphs \cite{Vignac2022,Tseng2023}, but the impact of different design choices in this setting has received little to no attention.

In this work, we explicitly explore how the underlying diffusion process may affect generative performance in a \textit{discrete-time} and \textit{discrete-object} setting. In particular, we will focus on generating \textit{undirected graphs}, as they are simply represented, yet arguably one of the most versatile and expressive discrete data types (i.e. many problems can be phrased as graph problems).

The space of possible discrete diffusion kernels is large. To simplify our analysis, we formulate a family of diffusion kernels based on the Bernoulli distribution, where only the noise schedule is a free parameter. We will show that adjusting the noise schedule induces a convergent prior distribution which---on graphs---is an Erd\"os--Renyi graph with any arbitrary edge probability $p$.

A few recent works have suggested intuitions for selecting the best convergent prior in a diffusion model. On continuous data types, \citet{Lee2021} achieved better performance on generating audio tracks with a diffusion prior which is a Gaussian with covariance equal to that of the original data distribution. For discrete diffusion on graphs, \citet{Vignac2022} proposed that the optimal prior should have the probability of each edge state (e.g. present or absent) match the empirical distribution in the original data. Together, these works have strongly suggested that different generation tasks merit the use of different diffusion priors, and they have intimated that the optimal prior is one whose core statistic (e.g. Gaussian covariance, multinomial probabilities, etc.) matches that of the original distribution \cite{Lee2021,Vignac2022}. We call this the \textit{empirical prior}. Importantly, although these works propose that the empirical prior is optimal, their results merely suggest that the empirical prior outperforms a uniform prior (e.g. isotropic Gaussian or uniform probabilities).

To our knowledge, this is the first work which \textit{systematically} explores how modifying the convergent prior directly affects generative performance in \textit{discrete} diffusion. Our results will challenge previous intuitions of what the optimal prior is. In particular, we highlight the following contributions:
\begin{itemize}
    \item We derive a novel family of discrete diffusion kernels based on asymmetric Bernoulli processes, that is easily adjustable so it converges to an arbitrary Erd\"os--Renyi prior.
    \item We demonstrate that different graph-generation tasks achieve optimal generative performance on diffusion kernels which converge to different priors.
    \item We show that the optimal prior for a given task is \textit{not} simply given by the empirical prior (i.e. based on statistics of the original data distribution) as previous works have suggested.
\end{itemize}

\section{An Adjustable, Asymmetric Bernoulli Kernel}
\label{sec:method}

Consider a bit $x_{t}$. At each time $t$, the diffusion process will flip the bit with probability according to some noise schedule. \citet{Tseng2023} proposed three such diffusion kernels, in which the final prior was a Bernoulli distribution of $\pi(x = 1) = 0,\pi(x = 1) = 1$, or $\pi(x = 1) = 0.5$. We extend from \citet{Tseng2023} by defining \textit{two} (potentially asymmetric) noise schedules: $\{\beta_{t}^{0},\beta_{t}^{1}\}$ for $t \in \{1,\cdots,T\}$. At time $t$, the bit $x_{t-1}$ is flipped to a 0 with probability $\beta_{t}^{0}$ (if $x_{t-1} = 1$), and is flipped to a 1 with probability $\beta_{t}^{1}$ (if $x_{t} = 0$). By defining these two distinct noise schedules, the final prior probability can be anything between 0 and 1. We will generally assume that $\beta_{t}^{b} \in [0, \frac{1}{2}]$.

We can derive the following forward-diffusion probability:
\begin{small}
\begin{equation}
\label{forward-eq}
q(x_{t} = 1\vert x_{0}) = \frac{1 + (-1)^{t-1}}{2} + \sum\limits_{i=1}^{t}\left[\frac{(-1)^{i}}{2}\epsilon_{i}^{\frac{1 + (-1)^{i}}{2}}\prod\limits_{j=i+1}^{t}\frac{\bar{\epsilon_{j}}}{2}\right] + x_{0}\prod\limits_{j=1}^{t}\frac{\bar{\epsilon_{j}}}{2}
\end{equation}
\end{small}

where $\epsilon_{t}^{b} = 2(1 - \beta_{t}^{b})$ and $\bar{\epsilon_{t}} = \epsilon_{t}^{0} + \epsilon_{t}^{1} - 2 = 2(1 - \beta_{t}^{0} - \beta_{t}^{1})$.

If $\lim\limits_{t\rightarrow T}\beta_{t}^{0} = p_{0}$ and $\lim\limits_{t\rightarrow T}\beta_{t}^{1} = p_{1}$ asymptotically, then the prior distribution is $q(x_{T} = 1) = \pi(x = 1) = \frac{p_{0}}{p_{0} + p_{1}}$. Thus, simply modifying the asymmetric noise schedules causes the diffusion process to converge to a Bernoulli distribution of any probability in the range $[0, 1]$ (Supplementary Figure \ref{fig:prior}). A full derivation of the kernel family is in Appendix \ref{kernel-deriv}.

In our work, we diffuse on graphs by treating the edges as binary states---either an edge exists or it does not. That is, for a graph of $n$ nodes, we diffuse over $\binom{n}{2}$ binary variables. We consider unlabeled nodes. Our adjustable Bernoulli kernel induces an Erd\"os--Renyi prior with probability $p = \frac{p_{0}}{p_{0} + p_{1}}$.

\section{Generative Performance Depends on the Prior}

We consider two well-known benchmark graph datasets: community (small) and stochastic block models. For each dataset, we trained discrete diffusion models using the adjustable Bernoulli kernel introduced in Section~\ref{sec:method}, exploring an extended range of prior probabilities corresponding to Erd\"os--Renyi graphs with $p$ in $\{0, 0.05, 0.10, ..., 0.95, 1\}$.

\begin{figure}[h]
\vskip 0.2in
\begin{center}
\centerline{\includegraphics[width=0.8\columnwidth]{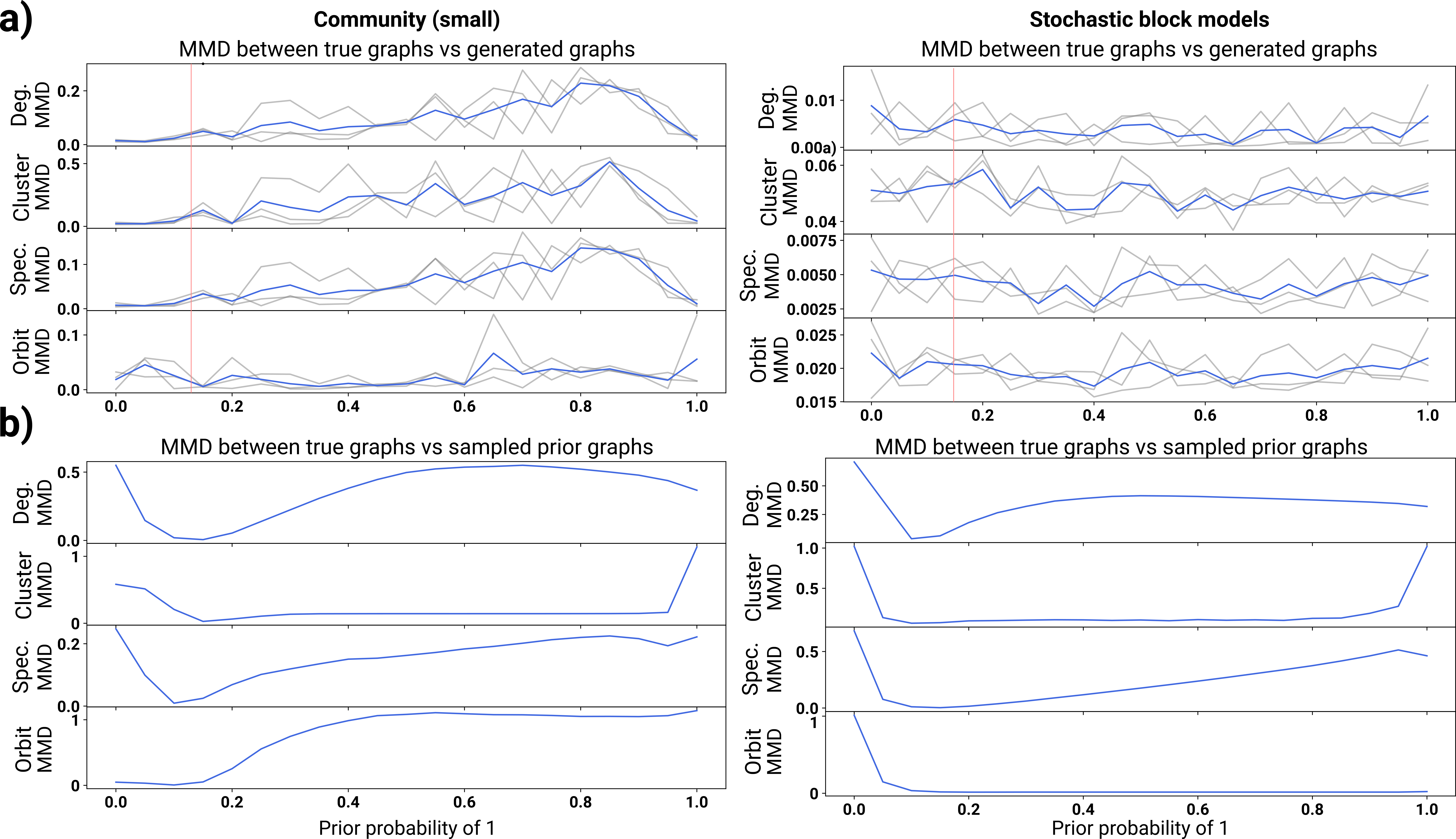}}
\caption{\textbf{a)} MMD of several graph distributions for our datasets, as a function of the prior probability in the diffusion kernel (the prior probability ranges from 0 to 1). A lower MMD is better. Three different random initializations are plotted in gray, and the average is in blue. The vertical red line marks the empirical probability of an edge in the original dataset. \textbf{b)} MMD between randomly sampled graphs from the prior distribution and the true data distribution, as a function of the prior probability.}
\label{fig:mmd}
\end{center}
\vskip -0.2in
\end{figure}

For each model, we quantified the generative performance by computing the maximum mean discrepancy (MMD) for several graph distributions, following previous works in the space of graph generation \cite{You2018,Liao2019,DeCao2018,Martinkus2022,Vignac2022}. This performance metric compares several distributions of various statistics over the generated and true graphs (i.e. distribution of node degrees, clustering coefficients, spectrum of the normalized Laplacian, and node orbit counts). Averaging over several random initializations, the MMD values show a clear \textit{preference} for which diffusion kernels---which vary in the convergent prior probability---yield the best performance overall (Figure \ref{fig:mmd}a). This preference is consistent regardless of which graph statistic MMD is computed on. Furthermore, the best kernel is \textit{different} between our datasets, and critically, the optimal kernel does not converge to the prior probability which matches the empirical probability in the original dataset. That is, the empirical prior is not necessarily optimal in our experiments. We also found that the generative performance of the optimal kernel yields better performance than previous graph-generative methods, including other discrete diffusion models (Supplementary Table \ref{table:performance-comp}).

It may seem intuitive to believe that the optimal prior should have a final edge probability that matches the empirical probability in the dataset (e.g. if the original dataset has a probability $p$ of having an edge, it may seem that the optimal diffusion kernel should also converge to a probability of $p$). In this regard, \citet{Lee2021} showed that a prior which matches the empirical data in covariance (in continuous Gaussian diffusion) could be learned by a simpler neural network (thus leading to a more efficient training). These same intuitions were in \citet{Vignac2022}, which showed some limited results for graphs suggesting that a diffusion kernel which converges to the empirical edge probability might have some moderate benefits over a uniform prior. Our experiments further extend these intuitions, and show that although the empirical prior may outperform the uniform prior, the optimal prior (at least for discrete graph diffusion) is not always the empirical prior.

\subsection{The Optimal Kernel is not Explained by Empirical MMD}

As the optimal diffusion kernel is not explained by the empirical distribution's edge probability, one may ask whether the kernel which yields the optimal MMD of generated graphs is the one whose prior distribution also has the optimal MMD (i.e. the one whose prior distribution matches the data distribution closest using MMD). In order to explore this, we sampled graphs from the prior distribution of each kernel, and computed the MMD between these randomly sampled graphs with the true data distribution.

Although we found that there was a trend in the convergent prior probability and the MMD between the prior distribution and the original data distribution (Figure \ref{fig:mmd}b), this optimum was \textit{not} the same as the optimal prior which maximizes generative performance (i.e. minimizes the MMD between the \textit{generated graphs} and the original data distribution). This optimum, however, \textit{does} match the empirical edge probability in the original data distribution.

\section{Searching for the Optimal Kernel in Practice}

Our results show that for discrete graph diffusion, the choice of diffusion prior can have large effects on the final generative performance. Additionally, the optimal prior is not simply the one which statistically matches the empirical data, or which maximizes similarity with the original data when measured by the MMD performance metric. Thus, we propose treating the diffusion kernel as a hyperparameter. In order to identify the optimal diffusion kernel, one may fix a family of diffusion kernels (e.g. Gaussian, or asymmetric Bernoulli as presented in Section~\ref{sec:method}, etc.) and search over it.

In order to aid in the efficient search for the optimal kernel, we found that by training only for a short time, the average training loss in the first few epochs is already somewhat predictive of the optimal kernel. That is, early training loss is correlated with final generative performance across different diffusion kernels in a family (Supplementary Figure \ref{fig:loss}). Furthermore, at least within the asymmetric-Bernoulli kernel family, we showed that the performance varies smoothly with the prior's probability of an edge (Figure \ref{fig:mmd}). This property is expected in other families of diffusion kernels (e.g. Gaussian kernels in continuous diffusion) and enables search through efficient hyper-optimization techniques, such as Bayesian optimization.


\section{Discussion}

In this work, we developed a family of diffusion kernels based on the Bernoulli distribution which is easily modified to tune the final prior probability of an edge. We demonstrated that the generative performance of a graph-generation task depends on the specific diffusion prior, and that the optimal kernel is different for different tasks. Critically, we showed how the optimal kernel is not defined by a prior whose underlying probability distribution is the same as the empirical probability distribution of the original data, as prior works have intuited. Instead, we suggested that the optimal kernel may be treated as a hyperparameter and tuned for, which can be done relatively efficiently.

Although the optimal kernel/prior was not obviously informed by the empirical data, our exploration paves the way for more research toward designing optimal priors for discrete diffusion models. Future work may explore potentially more inscrutable relationships which may explain the optimal kernel, as this remains an open problem in both discrete and continuous diffusion. 


\clearpage

\bibliography{branched_diffusion}

\begin{thebibliography}{15}
\providecommand{\natexlab}[1]{#1}
\providecommand{\url}[1]{\texttt{#1}}
\expandafter\ifx\csname urlstyle\endcsname\relax
  \providecommand{\doi}[1]{doi: #1}\else
  \providecommand{\doi}{doi: \begingroup \urlstyle{rm}\Url}\fi

\bibitem[Austin et~al.(2021)Austin, Johnson, Ho, Tarlow, and Berg]{Austin2021}
Austin, J., Johnson, D.~D., Ho, J., Tarlow, D., and Berg, R. V.~D.
\newblock Structured denoising diffusion models in discrete state-spaces.
\newblock \emph{Advances in Neural Information Processing Systems},
  22:\penalty0 17981--17993, 7 2021.
\newblock ISSN 10495258.
\newblock \doi{10.48550/arxiv.2107.03006}.
\newblock URL \url{https://arxiv.org/abs/2107.03006v2}.

\bibitem[Cao \& Kipf(2018)Cao and Kipf]{DeCao2018}
Cao, N.~D. and Kipf, T.
\newblock Molgan: An implicit generative model for small molecular graphs.
\newblock 5 2018.
\newblock \doi{10.48550/arxiv.1805.11973}.
\newblock URL \url{https://arxiv.org/abs/1805.11973v2}.

\bibitem[Dhariwal \& Nichol(2021)Dhariwal and Nichol]{Dhariwal2021}
Dhariwal, P. and Nichol, A.
\newblock Diffusion models beat gans on image synthesis.
\newblock \emph{Advances in Neural Information Processing Systems},
  11:\penalty0 8780--8794, 5 2021.
\newblock ISSN 10495258.
\newblock \doi{10.48550/arxiv.2105.05233}.
\newblock URL \url{https://arxiv.org/abs/2105.05233v4}.

\bibitem[Dockhorn et~al.(2021)Dockhorn, Vahdat, and Kreis]{Dockhorn2021}
Dockhorn, T., Vahdat, A., and Kreis, K.
\newblock Score-based generative modeling with critically-damped langevin
  diffusion.
\newblock 12 2021.
\newblock \doi{10.48550/arxiv.2112.07068}.
\newblock URL \url{https://arxiv.org/abs/2112.07068v4}.

\bibitem[Ho et~al.(2020)Ho, Jain, and Abbeel]{Ho2020}
Ho, J., Jain, A., and Abbeel, P.
\newblock Denoising diffusion probabilistic models.
\newblock \emph{Advances in Neural Information Processing Systems},
  2020-December, 6 2020.
\newblock ISSN 10495258.
\newblock \doi{10.48550/arxiv.2006.11239}.
\newblock URL \url{https://arxiv.org/abs/2006.11239v2}.

\bibitem[Karras et~al.(2022)Karras, Aittala, Aila, and Laine]{Karras2022}
Karras, T., Aittala, M., Aila, T., and Laine, S.
\newblock Elucidating the design space of diffusion-based generative models.
\newblock 6 2022.
\newblock \doi{10.48550/arxiv.2206.00364}.
\newblock URL \url{https://arxiv.org/abs/2206.00364v1}.

\bibitem[Lee et~al.(2021)Lee, Kim, Shin, Tan, Liu, Meng, Qin, Chen, Yoon, and
  Liu]{Lee2021}
Lee, S.-G., Kim, H., Shin, C., Tan, X., Liu, C., Meng, Q., Qin, T., Chen, W.,
  Yoon, S., and Liu, T.-Y.
\newblock Priorgrad: Improving conditional denoising diffusion models with
  data-dependent adaptive prior.
\newblock 6 2021.
\newblock URL \url{https://arxiv.org/abs/2106.06406v2}.

\bibitem[Liao et~al.(2019)Liao, Li, Song, Wang, Hamilton, Duvenaud, Urtasun,
  and Zemel]{Liao2019}
Liao, R., Li, Y., Song, Y., Wang, S., Hamilton, W.~L., Duvenaud, D., Urtasun,
  R., and Zemel, R.
\newblock Efficient graph generation with graph recurrent attention networks.
\newblock \emph{Advances in Neural Information Processing Systems}, 32, 10
  2019.
\newblock ISSN 10495258.
\newblock \doi{10.48550/arxiv.1910.00760}.
\newblock URL \url{https://arxiv.org/abs/1910.00760v3}.

\bibitem[Martinkus et~al.(2022)Martinkus, Loukas, Perraudin, and
  Wattenhofer]{Martinkus2022}
Martinkus, K., Loukas, A., Perraudin, N., and Wattenhofer, R.
\newblock Spectre : Spectral conditioning helps to overcome the expressivity
  limits of one-shot graph generators.
\newblock 4 2022.
\newblock \doi{10.48550/arxiv.2204.01613}.
\newblock URL \url{https://arxiv.org/abs/2204.01613v1}.

\bibitem[Rombach et~al.(2021)Rombach, Blattmann, Lorenz, Esser, and
  Ommer]{Rombach2021}
Rombach, R., Blattmann, A., Lorenz, D., Esser, P., and Ommer, B.
\newblock High-resolution image synthesis with latent diffusion models.
\newblock 12 2021.
\newblock \doi{10.48550/arxiv.2112.10752}.
\newblock URL \url{https://arxiv.org/abs/2112.10752v2}.

\bibitem[Sohl-Dickstein et~al.(2015)Sohl-Dickstein, Weiss, Maheswaranathan, and
  Ganguli]{Sohl-Dickstein2015}
Sohl-Dickstein, J., Weiss, E.~A., Maheswaranathan, N., and Ganguli, S.
\newblock Deep unsupervised learning using nonequilibrium thermodynamics.
\newblock \emph{32nd International Conference on Machine Learning, ICML 2015},
  3:\penalty0 2246--2255, 3 2015.
\newblock \doi{10.48550/arxiv.1503.03585}.
\newblock URL \url{https://arxiv.org/abs/1503.03585v8}.

\bibitem[Song et~al.(2021)Song, Sohl-Dickstein, Brain, Kingma, Kumar, Ermon,
  and Poole]{Song2021}
Song, Y., Sohl-Dickstein, J., Brain, G., Kingma, D.~P., Kumar, A., Ermon, S.,
  and Poole, B.
\newblock Score-based generative modeling through stochastic differential
  equations.
\newblock 2021.

\bibitem[Tseng et~al.(2023)Tseng, Diamant, Biancalani, and Scalia]{Tseng2023}
Tseng, A.~M., Diamant, N., Biancalani, T., and Scalia, G.
\newblock Graphguide: interpretable and controllable conditional graph
  generation with discrete bernoulli diffusion.
\newblock 2 2023.
\newblock URL \url{https://arxiv.org/abs/2302.03790v1}.

\bibitem[Vignac et~al.(2022)Vignac, Krawczuk, Siraudin, Wang, Cevher, and
  Frossard]{Vignac2022}
Vignac, C., Krawczuk, I., Siraudin, A., Wang, B., Cevher, V., and Frossard, P.
\newblock Digress: Discrete denoising diffusion for graph generation.
\newblock 9 2022.
\newblock \doi{10.48550/arxiv.2209.14734}.
\newblock URL \url{https://arxiv.org/abs/2209.14734v1}.

\bibitem[You et~al.(2018)You, Ying, Ren, Hamilton, and Leskovec]{You2018}
You, J., Ying, R., Ren, X., Hamilton, W.~L., and Leskovec, J.
\newblock Graphrnn: Generating realistic graphs with deep auto-regressive
  models.
\newblock \emph{35th International Conference on Machine Learning, ICML 2018},
  13:\penalty0 9072--9081, 2 2018.
\newblock \doi{10.48550/arxiv.1802.08773}.
\newblock URL \url{https://arxiv.org/abs/1802.08773v3}.

\end{thebibliography}
\bibliographystyle{icml2023}

\newpage
\appendix
\onecolumn

\section{Supplementary Figures and Tables}

\setcounter{figure}{0}
\renewcommand{\thefigure}{S\arabic{figure}}
\setcounter{table}{0}
\renewcommand{\thetable}{S\arabic{table}}

\begin{figure}[h]
\vskip 0.2in
\begin{center}
\centerline{\includegraphics[width=0.7\columnwidth]{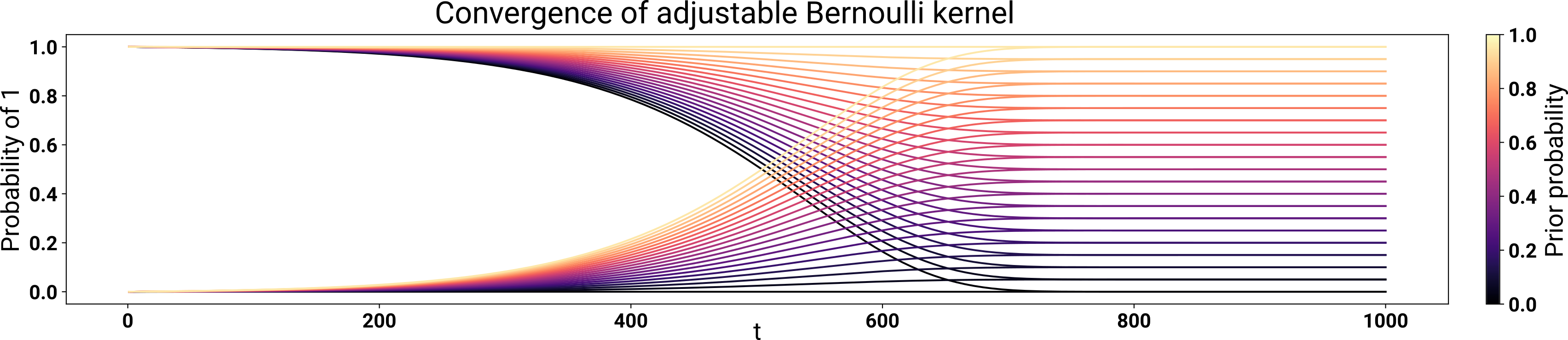}}
\caption{Visualization of the diffusion process of the adjustable Bernoulli kernel, for several different noise schedules. There are two lines of each color, showing the probability of a bit being 1 at each time $t$, if the original bit started at 0 or 1. Each color is a different asymmetric noise schedule, and the final probability converges to a prior defined by the asymptotic behavior of the noise schedules.}
\label{fig:prior}
\end{center}
\vskip -0.2in
\end{figure}

\begin{table}[h]
\caption{MMD ratio}
\label{table:performance-comp}
\begin{center}
\begin{tabular}{lcccccc}
Model & \multicolumn{3}{c}{Community (small)}  &  \multicolumn{3}{c}{Stochastic block models} \\
\midrule
 & Deg. $\downarrow$ & Clus. $\downarrow$ & Orbit $\downarrow$ & Deg. $\downarrow$ & Clus. $\downarrow$ & Orbit $\downarrow$ \\
\hline \\
GraphRNN & 2.00 & 1.31 & 2.00 & 2.62 & 1.33 & 1.75 \\
GRAN & 1.73 & 1.25 & 1.00 & 3.76 & 1.29 & 1.46 \\
MolGAN & 1.73 & 1.36 & 1.00 & 5.42 & 1.87 & 1.67 \\
SPECTRE & 1.00 & 1.73 & 1.00 & 3.14 & 1.26 & \textbf{0.54} \\
DiGress & 1.00 & 0.95 & 1.00 & 1.26 & 1.22 & 1.30 \\
\hline \\
Optimal prior & \textbf{0.99} & \textbf{0.57} & \textbf{0.79} & \textbf{0.56} & \textbf{1.18} & 0.83\\
\end{tabular}
\end{center}
\begin{center}
Comparison of generative performance of optimal prior to other works
\end{center}
\end{table}

\begin{figure}[h]
\vskip 0.2in
\begin{center}
\centerline{\includegraphics[width=0.5\columnwidth]{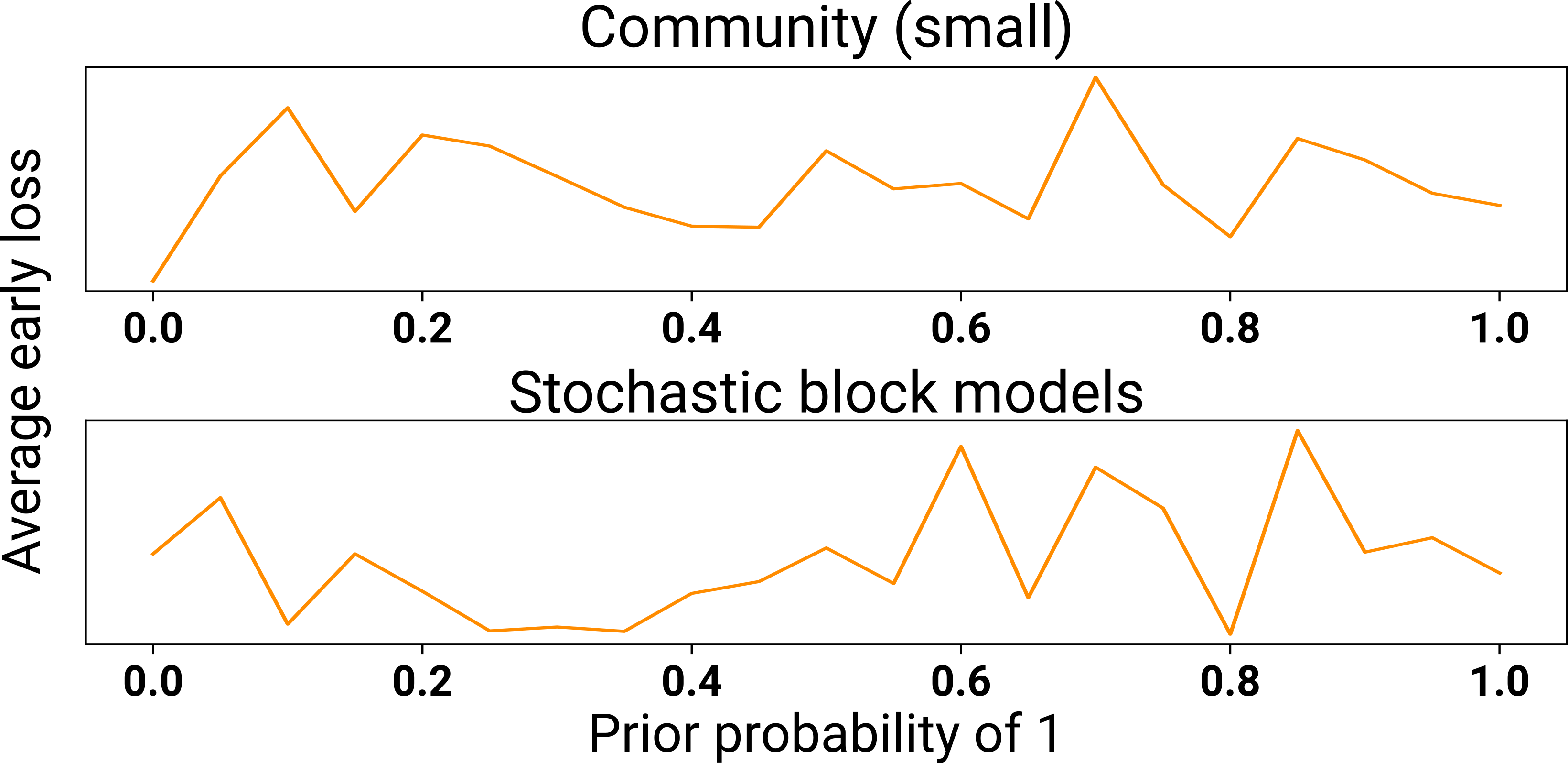}}
\caption{Average value of the loss for the first 10 epochs of training, for each diffusion kernel on each task.}
\label{fig:loss}
\end{center}
\vskip -0.2in
\end{figure}

\section{Derivation of asymmetric Bernoulli kernel}
\label{kernel-deriv}

\subsection{Forward diffusion distribution}

Here, we derive the forward distribution $q_{t}(x_{t} \vert x_{t-1}, x_{0})$. Note that every $x$ is a single bit.

Let us define a noising process $\{\beta_{t}^{0},\beta_{t}^{1}\}$ for $t \in \{1,\cdots,T\}$. In particular, we have $q(x_{t} = 1 \vert x_{t-1} = 0) = \beta_{t}^{0}$ and $q(x_{t}=0\vert x_{t-1}=1) = \beta_{t}^{1}$.

We will generally assume that $\beta_{t}^{b} \in [0, \frac{1}{2}]$.

In our derivation, we will use the following changes of variables to assist in simplification:

$\beta_{t}^{b} = 1 - \frac{1}{2}\epsilon_{t}^{b}$ (or equivalently, $\epsilon_{t}^{b} = 2(1 - \beta_{t}^{b})$)

$\bar{\epsilon_{t}} = \epsilon_{t}^{0} + \epsilon_{t}^{1} - 2 = 2(1 - \beta_{t}^{0} - \beta_{t}^{1})$

Below are the forward-distribution probabilities for the first four time steps:

$P(x_{1} = 1 \vert x_{0}) = \frac{1}{2}(2 - \epsilon_{1}^{0} + x_{0}\bar{\epsilon_{1}})$

$P(x_{2} = 1 \vert x_{0}) = \frac{1}{4}(2\epsilon_{2}^{1} - \epsilon_{1}^{0}\bar{\epsilon_{2}} + x_{0}\bar{\epsilon_{1}}\bar{\epsilon_{2}})$

$P(x_{3} = 1 \vert x_{0}) = \frac{1}{8}(8 - 4\epsilon_{3}^{0} + 2\epsilon_{2}^{1}\bar{\epsilon_{3}} - \epsilon_{1}^{0}\bar{\epsilon_{2}}\bar{\epsilon_{3}} + x_{0}\bar{\epsilon_{1}}\bar{\epsilon_{2}}\bar{\epsilon_{3}})$

$P(x_{4} = 1 \vert x_{0}) = \frac{1}{16}(8\epsilon_{4}^{1} - 4\epsilon_{3}^{0}\bar{\epsilon_{4}} + 2\epsilon_{2}^{1}\bar{\epsilon_{3}}\bar{\epsilon_{4}} - \epsilon_{1}^{0}\bar{\epsilon_{2}}\bar{\epsilon_{3}}\bar{\epsilon_{4}} + x_{0}\bar{\epsilon_{1}}\bar{\epsilon_{2}}\bar{\epsilon_{3}}\bar{\epsilon_{4}})$

Or in general:

$P(x_{t} = 1\vert x_{0}) = \frac{1}{2^{t}}(2^{t}(\frac{1 + (-1)^{t-1}}{2}) + \sum\limits_{i=1}^{t}[(-1)^{i}2^{i-1}\epsilon_{i}^{\frac{1 + (-1)^{i}}{2}}\prod\limits_{j=i+1}^{t}\bar{\epsilon_{j}}] + x_{0}\prod\limits_{j=1}^{t}\bar{\epsilon_{j}})$

In a more numerically stable form:

$P(x_{t} = 1\vert x_{0}) = \frac{1 + (-1)^{t-1}}{2} + \sum\limits_{i=1}^{t}[\frac{(-1)^{i}}{2}\epsilon_{i}^{\frac{1 + (-1)^{i}}{2}}\prod\limits_{j=i+1}^{t}\frac{\bar{\epsilon_{j}}}{2}] + x_{0}\prod\limits_{j=1}^{t}\frac{\bar{\epsilon_{j}}}{2}$

\subsection{Prior distribution}

By changing the value that $\beta_{t}^{0},\beta_{t}^{1}$ converge to, the prior can be made to be any probability between 0 and 1.

Now let us try and derive the prior probability more formally.

First, let us make the assumption that $T$ is even.

From above, we have that $P(x_{T} = 1\vert x_{0}) = x_{0}\bar{\epsilon}_{1}\cdots\bar{\epsilon}_{T} -\frac{1}{2^{T}}\epsilon_{1}^{0}\bar{\epsilon}_{2}\cdots\bar{\epsilon}_{T} + \frac{1}{2^{T-1}}\epsilon_{2}^{1}\bar{\epsilon}_{3}\cdots\bar{\epsilon}_{T} - \frac{1}{2^{T-2}}\epsilon_{3}^{0}\bar{\epsilon}_{4}\cdots\bar{\epsilon}_{T} + \cdots + \frac{1}{2}\epsilon_{T}^{1}$.

Early terms in this sequence consist of many $\bar{\epsilon}_{i}$ being multiplied together. For large $T$, these terms contribute an infinitesimal amount to the total sum. Thus, we can consider only the \textit{end behavior} of $\epsilon_{t}^{b}$. We make the simplifying assumption that $\beta_{t}^{0},\beta_{t}^{1}$ both approach some maximum value asymptotically, so the end behaviors of $\beta_{t}^{0},\beta_{t}^{1}$ are constant. This allows us to make the following substitutions for all times $t$ (as early times will contribute nothing to the final probability):

$\beta_{t}^{0}:=p_{0},\beta_{t}^{1}:=p_{1},\epsilon_{t}^{0}:=q_{0}=2(1-p_{0}),\epsilon_{t}^{1}:=q_{1}=2(1-p_{1}),\bar{\epsilon}_{t}:=s=2(1-p_{0}-p_{1})$ for all $t$

Then our expression becomes:

$P(x_{T} = 1\vert x_{0}) = -q_{0}\frac{1}{2^{T}}s^{T-2+1} + q_{1}\frac{1}{2^{T-1}}s^{T-3+1} - q_{0}\frac{1}{2^{T-2}}s^{T-4+1} + \cdots + q_{1}\frac{1}{2}$

We rearrange the terms by those with $q_{0}$ and those with $q_{1}$, and factor out $q_{0}$ and $q_{1}$ to obtain:

$P(x_{T} = 1\vert x_{0}) = -q_{0}\frac{s}{2^{2}}(1 + \frac{s^{2}}{2^{2}} + \frac{s^{4}}{2^{4}} + \cdots + \frac{s^{T-2}}{2^{T-2}}) + q_{1}\frac{1}{2}(1 + \frac{s^{2}}{2^{2}} + \frac{s^{4}}{2^{4}} + \cdots + \frac{s^{T-2}}{2^{T-2}})$

Now the series in the parentheses are geometric series. Recall, $\sum\limits_{i=0}^{n}r^{i} = \frac{1-r^{n-1}}{1-r}$. Thus, we get:

$P(x_{T} = 1\vert x_{0}) = -q_{0}\frac{s}{2^{2}}\sum\limits_{i=0}^{\frac{T}{2}-1}((\frac{s}{2})^{2})^{i} + q_{1}\frac{1}{2}\sum\limits_{i=0}^{\frac{T}{2}-1}((\frac{s}{2})^{2})^{i} = -q_{0}\frac{s}{2^{2}}\frac{1-(\frac{s}{2})^{T}}{1-(\frac{s}{2})^{2}} + q_{1}\frac{1}{2}\frac{1-(\frac{s}{2})^{T}}{1-(\frac{s}{2})^{2}}$

Now note that $(\frac{s}{2})^{T} \rightarrow 0$, so we get:

$P(x_{T} = 1\vert x_{0}) = -q_{0}\frac{s}{2^{2}}\frac{1}{1-(\frac{s}{2})^{2}} + q_{1}\frac{1}{2}\frac{1}{1-(\frac{s}{2})^{2}}$

Substituting back our original assumptions, we get:

$P(x_{T} = 1\vert x_{0}) = \frac{p_{0}}{p_{0} + p_{1}}$

Now let us consider the case where $T$ is odd.

From above, we have that $P(x_{T} = 1\vert x_{0}) = x_{0}\bar{\epsilon}_{1}\cdots\bar{\epsilon}_{T} -\frac{1}{2^{T}}\epsilon_{1}^{0}\bar{\epsilon}_{2}\cdots\bar{\epsilon}_{T} + \frac{1}{2^{T-1}}\epsilon_{2}^{1}\bar{\epsilon}_{3}\cdots\bar{\epsilon}_{T} - \frac{1}{2^{T-2}}\epsilon_{3}^{0}\bar{\epsilon}_{4}\cdots\bar{\epsilon}_{T} + \cdots - \frac{1}{2}\epsilon_{T}^{0} + 1$

We use the same assumptions as above for even $T$, and we obtain the following:

$P(x_{T} = 1\vert x_{0}) = -q_{0}\frac{1}{2}(1 + \frac{s^{2}}{2^{2}} + \frac{s^{4}}{2^{4}} + \cdots + \frac{s^{T-1}}{2^{T-1}}) + q_{1}\frac{1}{s}(\frac{s^{2}}{2^{2}} + \frac{s^{4}}{2^{4}} + \cdots + \frac{s^{T-1}}{2^{T-1}}) + 1$

Using the summation of a geometric series again, we get that $1 + \frac{s^{2}}{2^{2}} + \frac{s^{4}}{2^{4}} + \cdots + \frac{s^{T-1}}{2^{T-1}} = \frac{1-(\frac{s}{2})^{T+1}}{1-(\frac{s}{2})^{2}}$. Again, we can assume that $(\frac{s}{2})^{T+1}\rightarrow 0$.

Then $P(x_{T} = 1\vert x_{0}) = -q_{0}\frac{1}{2}(\frac{1}{1-(\frac{s}{2})^{2}}) + q_{1}\frac{1}{s}(\frac{1}{1-(\frac{s}{2})^{2}} - 1) + 1$

Substituting back our original assumptions, we get:

$P(x_{T} = 1\vert x_{0}) = \frac{p_{0}}{p_{0} + p_{1}}$ (the same as when $T$ is even)

\subsection{Posterior distribution}

We use Bayes’ Rule: $P(x_{t-1}=1\vert x_{t},x_{0}) = \frac{P(x_{t}\vert x_{t-1}=1,x_{0})P(x_{t-1}=1\vert x_{0})}{P(x_{t}\vert x_{0})}$.

We analyze each piece separately:

$P(x_{t}\vert x_{t-1}=1,x_{0})=x_{t}(1-\beta_{t}^{1}) + (1-x_{t})\beta_{t}^{1}$ (if $x_{t}=1$, this is the event we don’t flip from 1 to 0; if $x_{t}=0$, this is the event we do flip from 1 to 0).

$P(x_{t-1}=1\vert x_{0})$ comes directly from Equation \ref{forward-eq}.

$P(x_{t}\vert x_{0}) = x_{t}P(x_{t}=1\vert x_{0}) + (1-x_{t})(1-P(x_{t}=1\vert x_{0}))$, also from Equation \ref{forward-eq}.

This gives our posterior, $q_{t}(x_{t-1} \vert x_{t}, x_{0})$.

\end{document}